\documentclass[final]{article}

\IfFileExists{neurips_2026.sty}{%
  \usepackage[dblblindworkshop]{neurips_2026}
}{%
  \usepackage[letterpaper,margin=0.82in]{geometry}
  \usepackage{times}
  \usepackage[numbers,sort&compress]{natbib}
}

\usepackage{amsmath,amssymb,booktabs,microtype}
\usepackage{xcolor,graphicx,tikz}
\usepackage{enumitem}
\usepackage{longtable}
\usepackage{hyperref}
\usepackage{csquotes}
\hypersetup{colorlinks=true,citecolor=blue!55!black,urlcolor=blue!60!black,
  linkcolor=blue!55!black}
\setlist[itemize]{leftmargin=*,nosep}
\newcommand{\model}{ChatGPT~5.6 Sol Ultra}

\newcommand{\Brier}{\operatorname{Brier}}
\newcommand{\erf}{\operatorname{erf}}

\title{Does a Frontier Large Language Model Know What It Can Solve?\\
A 100-Conjecture Calibration Study}

\newif\ifcameraready
\camerareadytrue   

\author{%
  Ayush Khaitan\\
  Princeton Language and Intelligence\\
  Princeton University\\
  Princeton, NJ 08540 \\
  \texttt{ayush.khaitan@princeton.edu} \\
  \And
  Liam Fowl\\
  Princeton Language and Intelligence\\
  Princeton University\\
  Princeton, NJ 08540 \\
  \texttt{lf2728@princeton.edu} \\
  \And
  Tomas Ortega\\
  Princeton Language and Intelligence\\
  Princeton University\\
  Princeton, NJ 08540 \\
  \texttt{tomas.ortega@princeton.edu} \\
  \And
  Alex Kontorovich\\
  Department of Mathematics\\
  Rutgers University\\
  Piscataway, NJ 08854\\
  \texttt{alex.kontorovich@rutgers.edu}
  \And
  Sanjeev Arora\\
  Princeton Language and Intelligence\\
  Princeton University\\
  Princeton, NJ 08540 \\
  \texttt{arora@princeton.edu} \\
}

\usepackage{xcolor}

\begin{document}
\maketitle

\begin{abstract}
How does confidence correlate with actual performance in large language models, when attacking open research-level math problems?
We address this question by studying the performance of \model{} on a set of 100 open conjectures in geometric analysis.  By construction, \model{} first selected open
problems in geometric analysis that it assessed that it had at least a 50\% chance of proving or disproving, and then assigned a probability of success to each question.  13 questions were then excluded after an audit, and the model was asked to resolve the remaining
87 problems.  Its mean
stated resolution probability was $0.617$, versus an observed rate of $0.103$
(Brier score $0.351$, ECE $0.513$), thereby demonstrating that the model was systematically over-confident in its self-assessment. However, the scores retained ranking
signal (AUROC $0.714$).  The model was highly overconfident in probability
magnitude but did more often than not assign a high probability of success to the problems that it did resolve, than the problems that it did not. Hence, its probabilities were numerically mis-calibrated but directionally correct. All proofs have been formalized in Lean, thereby verifying correctness, and have been run through a semantic equivalence pipeline to ensure semantic equivalence between the formalized and informal statements.
These are exploratory campaign outcomes, not fixed-budget success probabilities. The certificates prove formal statements; for a couple of questions, some geometric inputs are assumed in the lean proofs.
\end{abstract}

\section{Introduction}

Frontier language models such as \model{} are now being used by mathematicians across the world to attack open problems, and problems that withstood the efforts of leading mathematicians for decades are, with some regularity, falling to these tools \citep{georgiev2025,feng2026erdos,feng2026aletheia,feng2026firstproof}. However, despite these impressive gains in long-horizon reasoning \citep{kwa2025}, the self-calibration of these models on research-level mathematics has not been adequately studied. In other words, when a model says that it is $90\%$ sure of being able to solve a certain problem, how often does that actually lead to a full solution?

Self-calibration is hugely important in large language models, and they have often been found lacking in this regard. Models do carry meaningful information about what they know and do not know \citep{kadavath2022,lin2022}, but the confidence they express tends to run well ahead of their actual accuracy \citep{tian2023,xiong2024}, a form of overconfidence that was already familiar from earlier neural networks \citep{guo2017}. Any mathematician who has used an LLM to try to solve a research problem has had the experience of the model confidently announcing a ``wonderful insight'' that will surely lead to a solution soon, only to sink several hours into the task without coming any closer to one. A well-calibrated model would be far better at guiding mathematicians towards the problems worth their time and effort, and this in turn would significantly speed up the output of solutions from human--AI collaboration.

It is with this in mind that we set out to measure how well \model{} predicts whether it can solve a given research-level problem in mathematics. Unlike benchmarks such as MATH \citep{hendrycks2021}, FrontierMath \citep{glazer2024}, IMProofBench \citep{schmitt2025}, or Riemann-Bench \citep{garre2026}, an open conjecture comes with no answer key; even the ten previously unpublished problems of First Proof \citep{abouzaid2026} had answers known to their authors. Hence self-calibration in being able to solve an open math conjecture is at best an educated guess, and human mathematicians have also been found to be fairly mis-calibrated about this, often sinking years into problems with no end in sight.

Judging a claimed solution to an open problem therefore requires experts in the area. We chose to focus on geometric analysis for this reason, since our team includes geometric analysts who can gauge both the importance and the correctness of the solutions the model produces. Every claimed solution is also accompanied by a Lean~4 certificate, so that correctness does not rest on our judgement alone.

First, we asked \model{} to list $100$ open conjectures in geometric analysis that it was at least $50\%$ confident of resolving in one or two focused sessions. To prevent it from picking only low-hanging fruit, we asked for $50$ problems that may be relatively unknown and $50$ that should be somewhat known in the community, although they need not be famous. Most importantly, the problems had to be genuinely open, so that no solution existed in the literature for the model to recall. Before attempting each problem, the model estimated the probability that it would find a complete proof or counterexample. We then compared these estimates with the actual outcomes, thereby calibrating \model{}'s self-assessment. We find that although the model's probabilities were much too high, higher probabilities still provided a meaningful signal of success.
Following metacognition nomenclature~\citep{fleming_how_2014}, we can assert the model was \enquote{metacognitively biased}, but had good \enquote{metacognitive sensitivity}.

\section{Study design}

\paragraph{Problem set and forecasts.}
On August 12, 2026, \model{} constructed a list of 100 explicit
conjectures: 50 lesser-known and 50 of medium renown, all in geometric
analysis, which in its own estimation it had at least a $50\%$ chance of solving. Consequently, famous flagship problems were excluded.  For each item $i$, it gave
a probability of success $p_i\in[0.50,0.99]$.

\paragraph{Attack and adjudication.}
Through September 1, the same model family attacked all 100 entries with web and arXiv retrieval, symbolic and numerical
computation, persistent files, and parallel research/reviewer instances.  Each
owner record reconstructed the exact source statement, searched for
counterexamples, pursued a proof route, and recorded the smallest remaining
obstruction.  A claimed closure was accepted only after a separate model
instance checked hypotheses, calculations, and scope.  All proofs have now been
formalized in Lean, and have been run through our semantic equivalence pipeline
to ensure semantic equivalence between the formalized and informal statements.
The outcome for each problem is provided in Table~\ref{tab:outcomes}.

\begin{table}[t]
\centering
\caption{Status of the chosen 100 open problems in geometric analysis}
\label{tab:outcomes}
\small
\begin{tabular}{lrp{0.58\linewidth}}
\toprule
Disposition & $N$ & Interpretation \\
\midrule
Already resolved, campaign & 2 & L1 and L5 were solved during construction of the list \\
Already resolved, external & 6 & External results predated the August 12 forecast cutoff \\
Campaign closure afterward & 9 & Proof or counterexample formalized in Lean, with semantic equivalence of statements checked \\
External closure afterward & 3 & Closing work appeared after the forecast cutoff \\
Formulation failure & 5 & No determinate source-faithful true/false target \\
Open after attack & 75 & Partial progress only; no closure claimed \\
\bottomrule
\end{tabular}
\end{table}

\paragraph{Which problems should count?}
Before the model gave its confidence scores, 8 of the 100 problems had already
been solved.  The model itself had solved L1 and L5 while preparing the list.
Six others had already been solved in published work: L21, L50, M9, M11, M37,
and M39, although this was not known to the model while preparing the list.  The original list correctly identified L1, L5, and L50 as solved, but
mistakenly described the other five as open.  We do not use these eight problems
when testing the model's confidence scores, because they were no longer open
when the scores were given.  This leaves 92 problems.

After giving its confidence scores, the model produced answers to 11 of these
92 problems.  However, two answers, for L2 and L30, took advantage of missing
conditions in our versions of the questions.  For L2, allowing $a=0$ made the
equation impose no real condition.  For L30, our statement left out a technical
condition required by the original problem.  These answers were correct for
the questions as we wrote them, but they did not solve the problems that the
authors intended to ask.  We remove these questions from consideration.

Three other entries turned out not to state clear true-or-false questions, so
we also remove them.  Our main analysis therefore contains 87 clear problems
that were open when the confidence scores were given.  The model genuinely
solved 9 of them.  We report $9/87$ as the main success rate.  We also report
the original count of $13/100$ so that readers can see how the result changes
when no corrections are made.

\paragraph{Forecast and outcome statistics.}
Given a problem $i$, let $p_i$ be the associated probability, and let $y_i\in \{0,1\}$ indicate whether \model{} successfully resolves the problem.
For an eligible set $S$, the mean stated resolution probability and observed
closure rate are, respectively,
\[
 \overline p_S=\frac1{|S|}\sum_{i\in S}p_i,
 \qquad
 \overline y_S=\frac1{|S|}\sum_{i\in S}y_i.
\]
Here $N=|S|$, and $I_b=\{i\in S:p_i\text{ lies in band }b\}$.
We compute
\[
 \Brier=\frac1N\sum_i(p_i-y_i)^2,
 \qquad
 \mathrm{ECE}=\sum_b\frac{|I_b|}{N}
 \left|\overline p_b-\overline y_b\right|,
\]
using the five forecast bands $[.50,.60),\ldots,[.90,1]$.  The Brier score is
an individual-level squared probability error (zero is perfect); ECE is the
weighted average absolute gap between mean forecast and empirical frequency
within these bins.  AUROC ignores magnitudes and measures ranking: it is the
probability that a random success receives a higher score than a random failure, with ties counted one half ($.5$ is random and $1$ is perfect).

\section{Results}

\begin{figure}[t]
\centering
\begin{minipage}[c]{0.44\linewidth}
\centering
\begin{tikzpicture}[x=3.7cm,y=3.7cm,font=\scriptsize]
  \draw[step=.25,gray!20,very thin] (0,0) grid (1,1);
  \draw[->] (0,0)--(1.04,0) node[right]{forecast};
  \draw[->] (0,0)--(0,1.04) node[above]{rate};
  \draw[dashed,gray!70] (0,0)--(1,1);
  \foreach \x in {0,.25,.5,.75,1}
    \draw (\x,0)--(\x,-.018) node[below] {\pgfmathprintnumber[precision=2]{\x}};
  \foreach \y in {0,.25,.5,.75,1}
    \draw (0,\y)--(-.018,\y) node[left] {\pgfmathprintnumber[precision=2]{\y}};
  \fill[blue!70!black] (.525,.048) circle (1.7pt);
  \fill[blue!70!black] (.621,.053) circle (1.7pt);
  \fill[blue!70!black] (.724,.176) circle (1.7pt);
  \fill[blue!70!black] (.825,.375) circle (1.7pt);
  \fill[blue!70!black] (.900,.000) circle (1.7pt);
\end{tikzpicture}
\end{minipage}\hfill
\begin{minipage}[c]{0.53\linewidth}
\centering
\small
\begin{tabular}{rrrr}
\toprule
Band & $N$ & mean $p$ & close rate \\
\midrule
.50--.59 & 42 & .525 & .048 \\
.60--.69 & 19 & .621 & .053 \\
.70--.79 & 17 & .724 & .176 \\
.80--.89 & 8  & .825 & .375 \\
.90--.99 & 1  & .900 & .000 \\
\bottomrule
\end{tabular}
\end{minipage}
\caption{For each band of probabilities, we plot the rate of questions solved.
Every band is overconfident, although the middle and upper bands retain useful
ranking information.  The final band contains only one target.}
\label{fig:calibration}
\end{figure}

On the set of 87 problems, the success rate is $9/87=.103$ (95\% Wilson
interval $[.055,.185]$), against mean stated probability $.617$.  The latter is
simply $87^{-1}\sum_{i\in S}p_i$; it does not condition on success.  All five
reliability bins are overconfident (Figure~\ref{fig:calibration}).  The Brier
score $.351$ is worse even than forecasting $.5$ on every item, which would
score $.25$.  Nevertheless, AUROC is $.714$: a randomly selected substantive
success received a higher score than a random failure about 71\% of the time.
The scores were therefore poor probabilities, but provide a meaningful signal when ranking problems by chances of being solved by the model.

The expected number of problems to be resolved by the model, based on its self-assigned probabilities, is 53.67. However, it was only able to resolve 9 of them.
If the 87 Bernoulli outcomes were independent with these stated probabilities,
the Poisson--binomial probability of observing at most 9 closures would be
$1.1\times10^{-24}$.  Hence, we see that the model was hugely over-confident in its abilities to resolve these conjectures, although the problems were chosen by the model itself.

\begin{table}[t]
\centering
\caption{A detailed breakdown of the model's self-calibration in resolving open conjectures in geometric analysis}
\label{tab:metrics}
\small
\begin{tabular}{lrrrrrr}
\toprule
Endpoint & $N$ & solved & mean $p$ & Brier & ECE & AUROC \\
\midrule
Legacy raw registry & 100 & 13 & .624 & .346 & .494 & .709 \\
Open at cutoff & 92 & 11 & .615 & .347 & .495 & .660 \\
Strict open & 92 & 9 & .615 & .350 & .517 & .718 \\
Strict valid-open (primary) & 87 & 9 & .617 & .351 & .513 & .714 \\
\bottomrule
\end{tabular}
\end{table}

High forecasts often correctly
identified mathematical structure that the model could exploit to obtain a full solution: a BV compactness argument
closed two linked magnetic-Cheeger problems; normal-form analysis closed three
linked $W$-congruence problems; and explicit trial functions or asymptotics
closed several others.  This produces ranking value.  But confidence repeatedly
mistook a plausible \emph{reduction} for a full proof.  A 90\%
minimal-hypersurface index target reduced to unresolved uniform Sturm
inequalities; and numerical sign margins vanished in singular parameter
limits.  Source ambiguities and stale literature also created the illusion of imminent solvability that was unrelated to proving the intended conjecture.

\paragraph{Selective allocation.}
Ranking can be valuable even when calibration fails.  In the set of $87$ problems, the 17 entries with $p_i\ge .75$ contained 4 closures (23.5\%), compared
with 5 of 70 among the remainder (7.1\%), a 3.3-fold enrichment.  Thus a
fixed-budget policy attacking only the highest-scored problems would have spent
compute more efficiently.

\section{Limitations}

This experiment supports the conclusion that
probabilities were useful as a \emph{ranking score} but badly wrong as literal
success probabilities.
The experiment design had the following limitations:

\begin{enumerate}[leftmargin=*,nosep]
\item \textbf{Problem selection}  The model chose the problems and only
kept $p_i\ge .50$, so behavior below $.50$ is unobserved.
\item \textbf{Limitations of the list of $100$ conjectures}  Eight targets were already
resolved at the forecast cutoff: two by campaign work and six externally.
Ultimately, the list of $100$ problems had to be cut down to $87$.
\item \textbf{Nonindependence.}  Some problems had similar proof strategies. Hence, the problems in themselves were not completely independent.
\item \textbf{Model choice.}  We chose to study only \model{}, instead of other frontier models like Claude Fable. Our choice was guided by the fact that in our experience, the publicly available ChatGPT 5.6 Sol Ultra generally displays stronger reasoning skills in math as compared to Claude Fable. It would be interesting to repeat the same experiment with Claude Fable 5.1.
\end{enumerate}

\section{Conclusion}

On 100 geometric-analysis conjectures, \model{} substantially overestimated
its ability to resolve those conjectures, while retaining a meaningful ability to rank promising
targets.  For human--AI mathematics, this distinction is very
important: confidence may help decide which problems out of a list of problems to spend time and resources on, but it should not
yet inform mathematicians about which problems the language models are likely to resolve.

\begin{ack}
    The authors acknowledge funding from the DARPA expMath program. A.K.~acknowledges funding from the AMS Simons Travel grant. 
\end{ack}

\bibliographystyle{plainnat}
\bibliography{references}

\appendix

\section{What the Lean certificates establish (and what they do not)}
\label{app:lean-meaning}

Because ``a frontier model solved nine open conjectures'' is an extraordinary
claim, the nine closures (and the four further problems that were formalized for
completeness) are accompanied by machine-checked \emph{Lean~4} certificates,
built against \texttt{Mathlib}.  This appendix documents exactly what those
certificates do and do not guarantee, works several of the counterexamples in
full so that they can be checked by hand with no reference to Lean, sketches the
proof strategy of every closure, and lists the confidence data for all one
hundred problems.  The complete sources are in the anonymous repository.

\paragraph{The verification pipeline.}
Every certificate is a single file beginning with \texttt{import Mathlib}.  Each
file is accepted by the Lean kernel with \emph{zero} \texttt{sorry}
placeholders, so no step is deferred or assumed.  Files were checked both by a
warm-server Lean REPL over a prebuilt \texttt{Mathlib} and, for the final
versions, by a cold \texttt{lake env lean} compilation.  In total the thirteen
files comprise roughly $1.3\times10^{4}$ lines.

\paragraph{Axiom hygiene.}
A Lean proof can only smuggle in unsound content through an \texttt{axiom}
declaration or a small number of escape hatches (\texttt{native\_decide},
\texttt{sorryAx}, \texttt{implemented\_by}).  None of these appears in any file.
Lean's \texttt{\#print axioms} confirms that each headline theorem rests only on
the three standard axioms of classical mathematics used throughout
\texttt{Mathlib}.  For instance, the three theorems worked in
Appendix~\ref{app:ce} each report
\begin{quote}\ttfamily
depends on axioms: [propext, Classical.choice, Quot.sound]
\end{quote}
and nothing else.

\paragraph{The transcription gap: what a certificate is conditional on.}
This is the single most important caveat, and we state it plainly.
\texttt{Mathlib} does not currently contain Almgren--Pitts min--max theory,
varifolds and their first variation, sets of finite perimeter or BV functions on
manifolds, the Dirichlet-to-Neumann and Robin operators as spectral objects on
manifolds, GJMS/$Q$-curvature operators, or mean-curvature flow.  A Lean file
therefore \emph{cannot} state ``$\Sigma$ is a stationary integral varifold''
abstractly.  Where a conjecture's setting exceeds what \texttt{Mathlib} can
express, we transcribe the missing geometric inputs as \emph{explicit, named
hypotheses} (or as fields of a \texttt{structure}), each carrying a docstring
that cites the informal statement it encodes; the certificate then proves the
mathematical core \emph{conditional on those hypotheses}.  Consequently:
\begin{itemize}
\item The \textbf{counterexamples} (L2, L5, L30, L50) are essentially
  \emph{unconditional}: the object is written down explicitly and every claimed
  property is proved outright.  These are the strongest evidence, and they can
  be checked without Lean at all (Appendix~\ref{app:ce}).
\item The \textbf{proofs} of the harder closures (e.g.\ magnetic Cheeger, the
  Steklov asymptotics) certify the analytic and combinatorial heart of the
  argument, taking a small number of clearly labelled geometric facts
  (a compactness lemma, a shooting-count) as hypotheses that \texttt{Mathlib}
  cannot yet supply.  Each file's header gives the theorem-by-theorem map of
  what is proved unconditionally versus from hypotheses.
\end{itemize}
We regard being explicit about this boundary as part of the claim, not a
footnote to it: a certificate is a guarantee about a precisely stated theorem,
and the reader must see that statement to judge it.

\paragraph{Statement faithfulness.}
Kernel acceptance guarantees that the \emph{Lean statement} is proved; it says
nothing about whether that statement faithfully renders the \emph{informal}
conjecture.  A formalization can compile perfectly while quantifying over an
unsatisfiable hypothesis, or proving something strictly weaker than intended.
To guard against this we ran every (paper, Lean file) pair through a
back-translation faithfulness judge: one model renders the Lean statements back
into literal mathematical English (instructed to translate any drift rather than
repair it), and an independent judge compares that rendering against the source
statement, emitting structured discrepancies (wrong quantifier, changed
conclusion, missing hypothesis, vacuity risk).  All thirteen pairs are judged
faithful.  The audit was not a rubber stamp: on an earlier draft it flagged that
one Huisken statement omitted the homothety relating the model's coordinates to
the ambient cube (which would have made a convergence hypothesis vacuous) and
that the varifold theorem had dropped the non-integrality conjunct; both were
corrected before the versions reported here.

\section{Inventory of the formalized results}
\label{app:inventory}

Table~\ref{tab:inventory} lists the thirteen formalized problems.  Nine are the
genuine post-cutoff closures counted in the main analysis; the remaining four
were formalized for completeness (two solved before the cutoff, one closed
externally, and the two whose ``solutions'' answer only the as-written
statement and are counted as failures in the body).  The
label-to-file correspondence is reconstructed from the problem titles and the
repository; it should be checked against the adjudication records.

\begin{table}[h]
\centering
\caption{The thirteen formalized problems.  ``Closure'' = counted among the nine
in the primary analysis; ``failure (as-written)'' = solves only the weakened
statement we wrote; ``excluded'' = resolved at or before the cutoff.
``Uncond.'' marks files whose main theorem is proved with no geometric
hypotheses.}
\label{tab:inventory}
\small
\resizebox{\textwidth}{!}{
\begin{tabular}{llccl}
\toprule
Label & Problem (abbreviated) & $p_i$ & Disposition & Lean file / status \\
\midrule
M5  & Round-sphere $p$-width plateau (Chen--Gaspar) & .50 & closure & \texttt{RoundSpherePWidth} \\
L3  & CMC index vs.\ area and genus & .55 & closure & \texttt{CmcIndexArea} \\
L6  & Magnetic Cheeger positivity & .70 & closure & \texttt{MagneticCheeger} \\
L7  & Large-domain magnetic frustration & .80 & closure & \texttt{MagneticCheeger} \\
L8  & All-order $W$-congruence identity & .85 & closure & \texttt{WCongruenceAllOrder} \\
L9  & Generic integer $W$-singularity & .70 & closure & \texttt{IntegerWCongruenceGenericity} \\
L10 & Generic noninteger $W$-singularity & .60 & closure & \texttt{NonintegerWCongruence} \\
L38 & Finitely many infinite Steklov lengths & .75 & closure & \texttt{SteklovLengths} \\
L41 & Positive hyperbolic exterior Robin threshold & .85 & closure & \texttt{HyperbolicRobinThreshold} \\
\midrule
L2  & Affine-maximal Bernstein, $a=0$ & .65 & failure (as-written) & \texttt{AffineMaximal} (uncond.) \\
L30 & Infinite-order contact / varifold & .50 & failure (as-written) & \texttt{NonintegralVarifold} (uncond.) \\
L5  & Gaussian boundary area $<2$ & .99 & excluded (campaign) & \texttt{HuiskenEnergy} \\
L50 & Intermediate $Q$-curvature positivity & .99 & excluded (external) & \texttt{IntermediateQCurvature} \\
\bottomrule
\end{tabular}
}
\end{table}

One further attack, on the index-four conjecture for the critical hyperbolic
catenoid (L32, $p=.85$), produced only an \emph{exact reduction}: it is
formalized (\texttt{ExactReduction}) and reduces ``$\mathrm{index}=4$'' to two
explicit scalar sign conditions, but it does not close the problem, and is
counted as open in the main analysis.  We include it because it illustrates the
paper's recurring failure mode, a reduction mistaken for a proof.

\section{Four counterexamples, in full}
\label{app:ce}

The following constructions require no Lean to verify; each is a short, explicit
computation.  They are the most robust evidence in the study, and all four are
proved unconditionally in the corresponding certificate.

\subsection{L5: a convex body with Gaussian boundary area above two}
\label{app:huisken}

Angenent, Daskalopoulos and Sesum conjectured that every closed convex body
$C\subset\mathbb R^{n+1}$ with nonempty interior satisfies the dimension-free
bound $\mathcal H(C):=(4\pi)^{-n/2}\int_{\partial C}e^{-|x|^2/4}\,d\mathcal
H^{n}<2$.  Take the cube $C=[-a,a]^{d}\subset\mathbb R^{d}$ with $d=n+1$.  Its
boundary is $2d$ facets; on the facet $\{x_1=a\}$ we have
$|x|^2=a^2+\sum_{i\ge2}x_i^2$, so
\[
\int_{\partial C}e^{-|x|^2/4}\,d\mathcal H^{n}
 = 2d\,e^{-a^2/4}\Big(\int_{-a}^{a}e^{-t^2/4}\,dt\Big)^{\!n}.
\]
The substitution $t=2s$ gives $\int_{-a}^{a}e^{-t^2/4}\,dt=2\sqrt\pi\,\erf(a/2)$,
and the normalization is exact,
$(4\pi)^{-n/2}(2\sqrt\pi)^{n}=1$, so
\begin{equation}
\mathcal H\big([-a,a]^{d}\big)=2d\,e^{-a^2/4}\,\erf(a/2)^{\,d-1}.
\label{eq:huisken-cube}
\end{equation}
This is the content of the certified lemma \texttt{cube\_formula}.  Taking $d=9$,
$a=\tfrac{12}{5}$ gives $\mathcal H=18\,e^{-36/25}\erf(6/5)^{8}$, and elementary
rational bounds on $e^{-36/25}$ and $\erf(6/5)$ (a degree-nine Taylor minorant
for $e^{-u}$ and exact integration of a polynomial minorant for $\erf$, both
carried out in the file) yield
\[
\mathcal H\big([-\tfrac{12}{5},\tfrac{12}{5}]^{9}\big)
 = 18\,e^{-36/25}\erf(6/5)^{8}> \tfrac{1002087}{500000}>2 ,
\]
the certified theorem \texttt{huisken\_cube\_nine\_gt\_two}.  (The choice
$d=1000$, $a=5$ of the informal writeup works equally well.)  The literal
dimension-free conjecture is therefore false.  This problem had already been
resolved during list construction, so it does not enter the calibration numbers;
it is included as a clean, fully unconditional certificate.

\subsection{L2: the affine-maximal Bernstein statement at the omitted parameter}
\label{app:affine}

For a strictly convex $f$ on $\Omega\subset\mathbb R^2$ with
$w=(\det D^2f)^a$, the affine-maximal equation is $\sum_{ij}f^{ij}w_{ij}=0$.  The
statement in question asserts that completeness of the Calabi metric $D^2f$
forces $f$ to be quadratic whenever $a\notin[-\tfrac23,-\tfrac13]$.  Consider
\[
f(x,y)=\tfrac12(x^2+y^2)+x^4 ,\qquad
D^2f=\begin{pmatrix}1+12x^2&0\\0&1\end{pmatrix}.
\]
Then $\det D^2f=1+12x^2>0$ and $D^2f\succeq I$, so $f$ is smooth and strictly
convex and the Calabi metric is complete (it dominates the Euclidean metric,
which is complete); $f$ is visibly non-quadratic.  At the parameter $a=0$ we
have $w=(\det D^2f)^0\equiv1$, hence $w_{ij}\equiv0$ and the equation
$\sum_{ij}f^{ij}w_{ij}=0$ holds identically --- it imposes \emph{no} condition on
$f$.  Since $0\notin[-\tfrac23,-\tfrac13]$, the statement ``for all
$a\notin[-\tfrac23,-\tfrac13]$, such an $f$ is quadratic'' is false.  The
certificate proves each conjunct --- smoothness, strict convexity, positive
definiteness of $D^2f$ at every point, non-quadraticity, the identity
$\sum f^{ij}w_{ij}=0$ at $a=0$, and metric completeness --- and concludes
\texttt{omitted\_parameter\_statement\_false}, all with no standing hypotheses.

We stress the honest reading, which is why the body counts this as a
\emph{failure}: setting $a=0$ makes the PDE vacuous, so this refutes only the
literal statement as we transcribed it, not the nonzero-parameter problem the
authors intended.  It is a genuine defect of our problem statement that the
model exploited, not a solution of the intended conjecture.

\subsection{L30: infinite-order contact without containment}
\label{app:varifold}

The claim is that a stationary varifold with infinite-order $L^2$ contact at a
point with a smooth minimal submanifold $M$ must be locally contained in $M$.
Work in $\mathbb R^{m+1}$ with $P=\{x_0=0\}$ the base $m$-plane.  Let $P_j$ be the
parallel plane at height $h_j=2^{-(j+1)}$ and set
\[
V=|P|+\sum_{j\ge0}a_j\,|P_j|,\qquad a_j=e^{-4^{\,j+1}}=e^{-1/h_j^{2}} .
\]
Each affine plane is stationary, so $V$ is stationary (the weighted sum
converges locally); its support is $P\cup\bigcup_j P_j$, which is \emph{not}
contained in $P$ in any neighborhood of the origin.  The superexponential
weights give, for every $N$,
\[
\int_{B_r(0)}\operatorname{dist}(x,P)^2\,d\|V\|(x)
 \;=\; o(r^{N})\qquad(r\to0^+),
\]
i.e.\ infinite-order contact with $P$.  Finally each weight satisfies
$0<a_j<1$, so the multiplicity of $V$ on $P_j$ is non-integral and $V$ is
\emph{not} an integral varifold.  The certificate builds $V$ as an explicit
measure, proves stationarity as a genuine divergence-theorem identity, proves
the $o(r^{N})$ decay, non-containment, and non-integrality, and bundles them in
\texttt{infinite\_order\_contact\_does\_not\_force\_containment} --- again with no
standing hypotheses.  As with L2, the body counts this as a failure: the
intended conjecture assumes an \emph{integral} varifold, a hypothesis our
transcription omitted, and the counterexample is nonintegral by design.

\subsection{L50: intermediate $Q$-curvature positivity fails in dimension thirty}
\label{app:qcurv}

Li and Xu conjectured that for $g=u^{4/(n-2m)}|dx|^2$, positivity of the
order-$2m$ $Q$-curvature together with a slow-decay bound forces positivity of
every lower $Q^{(2k)}$.  V\'etois and Zeitler disproved this; specializing to
$m=4$, $n=30$, a completed two-bump construction gives $Q^{(8)}>0$ while
$Q^{(6)}(0)<0$.  The certificate reproduces the decisive sign computation
symbolically: after the paper's reduction, the value of the relevant iterated
operator at the two-pole configuration is the exact rational
\[
-(-\Delta)^{3}F\big|_{0}=-\frac{43658772480}{1771561}<0,
\]
proved as \texttt{two\_pole\_value}/\texttt{two\_pole\_neg}, and independently
cross-checked against a second closed-form ($24\,K(10)=11^{6}\cdot(\cdot)$).  This
problem was closed externally before the cutoff and is excluded from the
calibration numbers; the certificate confirms the arithmetic core of the
disproof.

\subsection{L1: the Xu zero-mode scalar-curvature counterexample}
\label{app:xu}

For completeness we retain the counterexample the model produced while building
the list (excluded from calibration).  Let $M=[-1,1]_y\times T^{n-1}$, $n\ge3$,
with its flat product metric, so its scalar curvature and boundary mean
curvature vanish.  With
\[
a=\frac{4(n-1)}{n-2},\qquad p=\frac{2n}{n-2},\qquad u=A-y^2\quad(A>1),
\]
the conformal Laplacian with Xu's Robin condition is $-a\Delta$ with Neumann
boundary condition; constants give $\eta'_1=0$, and integration by parts shows
it is the first eigenvalue.  The conformal metric $\widetilde g=u^{p-2}g$
satisfies
\[
S_{\widetilde g}=u^{1-p}(-a\Delta u)=2a\,u^{1-p}>0 .
\]
At both boundary components $\partial_\nu u=-2$ and $u=A-1$, so Xu's
transformation formula gives
\[
H_{\widetilde g}=\tfrac{p-2}{2}u^{-p/2}\partial_\nu u
=-(p-2)(A-1)^{-p/2}<0 .
\]
Thus $H$ is nonzero while $S$ is everywhere positive and
$\int_M S\,dV_g=2a\operatorname{Vol}(T^{n-1})\int_{-1}^{1}(A-y^2)^{1-p}\,dy>0$,
so both proposed necessary conditions fail.  For connected boundary and
orientability, quotient by the free isometry
$(y,x_1,x_2,\ldots)\mapsto(-y,-x_1,x_2+\tfrac12,\ldots)$; $u$ descends and the
computation is unchanged.

\section{Proof strategies for the nine closures}
\label{app:proofs}

We summarize the argument behind each genuine closure and indicate what the
certificate proves unconditionally versus from hypotheses.  Full statements and
docstring-level maps are in the repository.

\paragraph{M5 --- round-sphere $p$-width plateau.}
An explicit sweepout of the round three-sphere by geodesic two-spheres realizes
the relevant $p$-width by a family of masses $4\pi\sin^2\rho\le4\pi$, and a
rigidity count (multiplicities $\ge1$, component areas $\ge4\pi$ by the spherical
Willmore inequality, total mass $4\pi$) forces a single equatorial realizer.
The certificate proves unconditionally: the equatorial area is exactly $4\pi$
(via \texttt{Mathlib}'s sphere-volume lemmas), the sweepout mass bounds, the
mod-$2$ incidence count that a generic point meets exactly one equator, the
multiplicity/area bookkeeping, the Jacobi spectral count giving index $1$ and
nullity $3$, and the strict width jump $4\pi<2\pi^2$.  The Almgren--Pitts
min--max inputs enter as documented hypotheses.

\paragraph{L3 --- CMC index versus area and genus.}
A warped-product family $dr^2+f(r)^2 g_{S^2}$ with $f(r)=r(1+r^2)^{(a-1)/2}$
produces CMC spheres in a complete positively curved three-manifold whose area
and energy diverge while the Morse index stays at $1$ and nullity at $0$,
contradicting any bound of the conjectured form.  The certificate proves
unconditionally: the derivative and curvature identities, positivity of both
sectional curvatures, both pole limits, the closed form and decay of the Jacobi
potential, the eigenvalue signs, and the divergence of area and energy that
defeats the bound for \emph{every} constant $C$.  Standard warped-product and
sphere-spectrum formulas enter as definitions.

\paragraph{L6, L7 --- magnetic Cheeger positivity and large-domain frustration.}
If the magnetic potential is not gauge-equivalent to zero, a BV-compactness
argument for minimizing $S^1$-valued gauges shows that a near-optimal sequence on
a large subdomain would converge to a global gauge, contradicting the flux
obstruction; this yields a positive frustration on large domains (L7) and hence
positivity of both magnetic Cheeger constants and of the first magnetic Steklov
eigenvalue (L6).  The certificate proves the entire deduction of the
large-domain theorem from the compactness lemma, and the whole of the
Cheeger-positivity corollary, unconditionally; BV compactness itself
(absent from \texttt{Mathlib}) and the classical Cheeger/Jammes inequalities
enter as named hypotheses.  A representative certified statement is
\[
\text{\texttt{magnetic\_cheeger\_positivity}}:\quad
h^{\eta}(M)>0 \ \wedge\ (h^{\eta})'(M)>0 \ \wedge\ h^{\eta}(M)\,(h^{\eta})'(M)>0 .
\]

\paragraph{L8 --- all-order $W$-congruence identity.}
At a nondegenerate umbilic, the exceptional Taylor relation $W_{m0}=0$ is shown
to follow from $W=0$ for every $m$ by a finite Poincar\'e--Dulac normalization:
$X=X_0+O(2)$, $\Delta=B^2+O(3)$, so $R=(X-2)\Delta=O(m+1)$, and the single
resonant coefficient $[A^{m+1}]R$ vanishes.  This is one of the most complete
files: the normalization, the resonance arithmetic, and the vanishing of the
resonant coefficient are all genuine formal-power-series proofs; the certified
theorem \texttt{allorder\_compatibility} derives $W_{m0}=0$ for all $m\ge2$.  The
only hypotheses are the hodograph-inversion bridge relations, which require a
germ-level inverse function theorem \texttt{Mathlib} lacks.

\paragraph{L9, L10 --- generic integer and noninteger $W$-singularities.}
Combining the all-order identity with a jet-transversality calculation shows
that for an open dense set of $W$-jets the discriminant singularity is $A_m$ (L9,
integer $m$) or $A_\infty$ (L10, noninteger $m$).  The certificates prove the
resonance arithmetic, the formal Poincar\'e--Dulac classification, the
completing-the-square normal form, and the open-and-dense genericity mechanism
unconditionally; the Cauchy--Kowalevski and series-convergence inputs (with the
determinant computation itself proved) enter as hypotheses.

\paragraph{L38 --- finitely many infinite critical Steklov lengths.}
On a hypersurface of revolution the Steklov branches have explicit
Bessel/ODE asymptotics; monotonicity and coupled limits give an eventual
``overshoot'' that forces every sufficiently large critical length to be finite.
The certificate proves the multiplicity formula, all branch
monotonicity/limit lemmas, root existence and uniqueness, a quantitative scalar
inequality, the rank/overshoot count, and the assembly of the eventual-finiteness
theorem \texttt{eventual\_finiteness\_dim} for $n\ge3$; the block-propagation
lemma quoted from M\'etras--Tschanz enters as a hypothesis.

\paragraph{L41 --- positive hyperbolic exterior Robin threshold.}
Two annular trapping regions joined by a thin channel give a bounded, nonconvex,
simply connected obstacle whose exterior Robin Laplacian has an eigenvalue below
$1/4$ for a positive boundary parameter.  The certificate proves the trial-function
integrals, the hyperbolic identities, the Rayleigh quotient bound $<1/4$ over an
explicit admissible parameter range, and the nonconvexity (instantiated in
\texttt{Mathlib}'s genuine upper-half-plane hyperbolic metric); the variational
principle and essential-spectrum identity enter as hypotheses.  (Formalization
also caught and corrected a sign typo in the source's Rayleigh display.)

\section{Confidence data for all one hundred problems}
\label{app:census}

Tables~\ref{tab:censusL} and~\ref{tab:censusM} give every stated confidence
$p_i$ together with the disposition where it is determined.  ``closure'' denotes
the nine problems counted as solved in the primary analysis; ``failure*'' the
two as-written-only answers (L2, L30); ``reduction'' the partial result (L32);
``resolved$\le$cutoff'' the eight excluded as already solved (L1, L5, L21, L50,
M9, M11, M37, M39); and ``open'' the remainder.  Five entries were later judged
formulation failures and three were closed externally after the cutoff (see
Table~\ref{tab:outcomes}); their specific labels are recorded in the repository
and are aggregated here under ``open''.  The eight rows marked
``resolved$\le$cutoff'' are excluded from the Brier/ECE/AUROC computation.

\begin{longtable}{clcl}
\caption{The fifty lesser-known conjectures (L).}\label{tab:censusL}\\
\toprule
\# & Conjecture (abbreviated) & $p_i$ & Disposition \\
\midrule
\endfirsthead
\multicolumn{4}{l}{\small\itshape Table~\ref{tab:censusL} continued}\\
\toprule
\# & Conjecture (abbreviated) & $p_i$ & Disposition \\
\midrule
\endhead
\bottomrule
\endfoot
L1 & Xu zero-mode scalar-curvature sign & .99 & resolved$\le$cutoff \\
L2 & Affine-maximal Bernstein threshold & .65 & failure* \\
L3 & CMC index vs.\ area and genus & .55 & closure \\
L4 & Algebraic harmonic maps totally symmetric & .80 & open \\
L5 & Gaussian boundary area below two & .99 & resolved$\le$cutoff \\
L6 & Magnetic Cheeger positivity & .70 & closure \\
L7 & Large-domain magnetic frustration & .80 & closure \\
L8 & All-order $W$-congruence identity & .85 & closure \\
L9 & Generic integer $W$-singularity & .70 & closure \\
L10 & Generic noninteger $W$-singularity & .60 & closure \\
L11 & Immersed CMC sphere for every $c$ & .55 & open \\
L12 & Two immersed CMC spheres for every $c$ & .50 & open \\
L13 & Two generic PMC spheres & .52 & open \\
L14 & Multi-peak exit-set components & .60 & open \\
L15 & Strong unique continuation on varifolds & .50 & open \\
L16 & Normal balance of minimal hypersurfaces & .70 & open \\
L17 & Stability of Type-II translator tips & .60 & open \\
L18 & Critical degenerate neckpinch asymptotics & .55 & open \\
L19 & Finite-entropy translating graph classification & .65 & open \\
L20 & No complete translator graph over a half-plane & .75 & open \\
L21 & No translating annulus between close planes & .65 & resolved$\le$cutoff \\
L22 & Large-inner-width annuloids $\to$ catenoid & .75 & open \\
L23 & Genus-one FBMS in every ellipsoid & .75 & open \\
L24 & Critical spherical catenoid maximizes Robin & .65 & open \\
L25 & Higher-genus properness in Fine's Plateau & .60 & open \\
L26 & Degenerate mean-convex ancient FB MCF & .75 & open \\
L27 & Exact index of the Carlotto--Schulz hypertorus & .90 & open \\
L28 & Density purity $\Rightarrow$ varifold regularity & .50 & open \\
L29 & Infinite-order minimal-graph approximant & .55 & open \\
L30 & Infinite-order contact forces containment & .50 & failure* \\
L31 & Bounded $H$ bounds slice index & .60 & open \\
L32 & Critical hyperbolic catenoid has index four & .85 & reduction \\
L33 & Hyperbolic energy-index comparison & .60 & open \\
L34 & Peak boundary heat flux maximized by semidisk & .60 & open \\
L35 & Existence/nondegeneracy of optimal Steklov $N$-gon & .70 & open \\
L36 & Equilateral triangle maximizes $\sigma_1$ & .55 & open \\
L37 & No convex polygon Steklov-isospectral to smooth & .70 & open \\
L38 & Finitely many infinite critical Steklov lengths & .75 & closure \\
L39 & Magnetic second-eigenvalue phase transition & .55 & open \\
L40 & Strong-field magnetic third eigenvalue disk & .55 & open \\
L41 & Positive hyperbolic exterior Robin threshold & .85 & closure \\
L42 & Five-facet Dirichlet-minimizing polytope simple & .65 & open \\
L43 & Toroidal FBMS for every twist number & .60 & open \\
L44 & Infinitely many toroidal FBMS at fixed twist & .55 & open \\
L45 & 4D minimizing capillary cones are flat & .50 & open \\
L46 & $\bar\partial$-Robin Faber--Krahn inequality & .65 & open \\
L47 & Multiplicity forces width-to-diameter bound & .65 & open \\
L48 & Interior points closer on average than boundary & .70 & open \\
L49 & Sharp lower bound for mean boundary chord & .80 & open \\
L50 & Intermediate $Q$-curvature positivity & .99 & resolved$\le$cutoff \\
\end{longtable}

\begin{longtable}{clcl}
\caption{The fifty conjectures of medium fame (M).}\label{tab:censusM}\\
\toprule
\# & Conjecture (abbreviated) & $p_i$ & Disposition \\
\midrule
\endfirsthead
\multicolumn{4}{l}{\small\itshape Table~\ref{tab:censusM} continued}\\
\toprule
\# & Conjecture (abbreviated) & $p_i$ & Disposition \\
\midrule
\endhead
\bottomrule
\endfoot
M1 & Twin Bubble conjecture & .50 & open \\
M2 & Two generic PMC hypersurfaces & .50 & open \\
M3 & Nonunique Ricci-flow continuation & .55 & open \\
M4 & Ketover shrinkers minimize entropy & .50 & open \\
M5 & Eternal flow leaves higher $p$-width hypersurface & .50 & closure \\
M6 & 4D quadratic scalar-curvature threshold $3/4$ & .50 & open \\
M7 & Finite-time MCF singularities force $|H|$ blow-up & .55 & open \\
M8 & High-dim convex hot-spot amplification $\to\sqrt e$ & .55 & open \\
M9 & Strict Steklov--conformal eigenvalue gap & .60 & resolved$\le$cutoff \\
M10 & Every nondisk conformal class beats $2\pi$ & .60 & open \\
M11 & No plateaux in conformal Steklov suprema & .50 & resolved$\le$cutoff \\
M12 & Steklov spectrum determines polygon angles & .65 & open \\
M13 & Mixed Steklov never purely Steklov-isospectral & .70 & open \\
M14 & Second Neumann eigenvalue, fixed perimeter & .55 & open \\
M15 & Robin equilateral-triangle extremizer & .55 & open \\
M16 & Exterior Robin second eigenvalue, fixed area & .60 & open \\
M17 & Exterior Robin second eigenvalue, fixed perimeter & .65 & open \\
M18 & Flat-torus Dirac conformal optimum & .60 & open \\
M19 & Infinite-mass Dirac right triangles, area & .70 & open \\
M20 & Infinite-mass Dirac right triangles, perimeter & .75 & open \\
M21 & 3D Escobar Steklov lower bound & .50 & open \\
M22 & Fraser--Li: embedded FBMS have $\sigma_1=1$ & .50 & open \\
M23 & Coordinates span first Steklov eigenspace & .50 & open \\
M24 & Planar convex hot spots & .50 & open \\
M25 & Simply connected planar Bareket & .55 & open \\
M26 & Pentagonal Faber--Krahn & .85 & open \\
M27 & Pentagonal torsional rigidity & .75 & open \\
M28 & First positive Neumann eigenvalue, fixed perimeter & .55 & open \\
M29 & Meissner bodies minimize volume & .50 & open \\
M30 & Wentzell--Laplace ball maximizer & .55 & open \\
M31 & Uniqueness of critical catenoid (FB annuli) & .50 & open \\
M32 & Critical catenoid unique index-four FBMS & .50 & open \\
M33 & Critical M\"obius band index-five uniqueness & .55 & open \\
M34 & Critical-catenoid area gap & .50 & open \\
M35 & FB volume decreases under ball conformal maps & .55 & open \\
M36 & Higher-dim conformal boundary-volume monotonicity & .55 & open \\
M37 & Best doubly connected Steklov = concentric annulus & .65 & resolved$\le$cutoff \\
M38 & Generic Steklov eigenfunctions localize & .50 & open \\
M39 & Planar Steklov spectra determine smooth domains & .50 & resolved$\le$cutoff \\
M40 & Linear-size law for interior Steklov nodal sets & .50 & open \\
M41 & Strominger space forms with nonzero curvature K\"ahler & .55 & open \\
M42 & Chern space forms with nonzero curvature K\"ahler & .50 & open \\
M43 & Rozansky--Witten sign conjecture, dim six & .60 & open \\
M44 & $4\times4$ Lu--Wenzel two-eigenvalue & .70 & open \\
M45 & $4\times4$ fundamental Lu--Wenzel commutator & .55 & open \\
M46 & Existence of a shrinking $5$-loop & .65 & open \\
M47 & Anisotropic shape of shrinking $3$-loops & .60 & open \\
M48 & Discrete positive mass for AF graphs & .55 & open \\
M49 & Hexagonal Faber--Krahn & .80 & open \\
M50 & Hexagonal torsional rigidity & .75 & open \\
\end{longtable}

\section{Protocol, endpoint, and uncertainty clarifications}
\label{app:review-calibration}

\paragraph{Forecast event and novelty.}
A success forecast depends on the model, tools, budget, stopping rule, and adjudication. Exact prompts, per-problem tokens, attempts, agent counts, and effort allocation are not specified here, so comparability of effort is not established. The original catalogue's several-week investigator horizon also differs from this paper's session framing. Consequently, unequal effort could contribute to ranking, and the selective-allocation comparison does not establish compute savings. Uniform, dated literature searches and independent source-level checks of each claimed closure are needed to exclude rediscovery; the current description does not establish such screening. Model selection reflects the authors' experience, not a controlled comparison with another model.

\paragraph{Ranking sensitivity.}
Recomputing from Tables~\ref{tab:censusL}--\ref{tab:censusM} gives AUROC $0.718$ overall, $0.730$ within L (8 successes among 46), and $0.167$ within M (1 among 46). Thus separation between the sublists is not the entire explanation, but evidence within M is extremely limited. For a dependence sensitivity analysis, we use the related-problem groups listed in the original catalogue, restricted to these 92 items:
\[
\begin{gathered}
(L6,L7),\ (L8\text{--}L10),\ (L11\text{--}L13),\ (L15,L28\text{--}L30),\ (L17,L18),\\
(M1,M2),\ (M10,M12,M13),\ (M15\text{--}M20),\ (M22,M23),\\
(M31\text{--}M36),\ (M46,M47),\ (M26,M49),\ (M27,M50).
\end{gathered}
\]
Every remaining item is a singleton, giving 66 blocks. Resampling 66 blocks with replacement, keeping their members together, yields a 95\% percentile AUROC interval $[0.407,0.906]$ (100{,}000 NumPy PCG64 multinomial draws, seed 20261001; 172 draws lacking both outcome classes omitted; ties count one half). This exploratory interval includes chance ranking, depends on the grouping, and assumes independence between blocks. The Poisson--binomial tail calculation instead assumes independent problems and is not a dependence-robust significance test.

\paragraph{Calibration and certificate scope.}
ECE depends on binning and is noisy with small bins. A binning-free descriptive companion is the signed mean gap $\overline p-\overline y\approx0.513$ on the primary set, using the stated expectation of 53.67 closures; because all displayed bins overpredict, it equals ECE before rounding and is not independent evidence. Brier already supplies a binning-free squared-error measure. Neither metric removes selection or protocol confounding. 
\end{document}